\documentclass[10pt,twocolumn,letterpaper]{article}

\usepackage{cvpr}              

\usepackage{graphicx}
\usepackage{amsmath}
\usepackage{amssymb}
\usepackage{booktabs}
\usepackage{microtype}      
\usepackage{xcolor}         
\usepackage{graphicx}
\usepackage{multirow}
\usepackage{bm}
\usepackage{caption}

\definecolor{cvprblue}{rgb}{0.21,0.49,0.74}
\usepackage[pagebackref,breaklinks,colorlinks,citecolor=cvprblue]{hyperref}

\def\paperID{6185} 
\def\confName{CVPR}
\def\confYear{2024}

\title{Active Open-Vocabulary Recognition: Let Intelligent Moving\\Mitigate CLIP Limitations}

\author{
Lei Fan,
Jianxiong Zhou, 
Xiaoying Xing and Ying Wu\\
Northwestern University\\
{\tt\small \{leifan,jianxiongzhou2026,xiaoyingxing2026\}@u.northwestern.edu,yingwu@northwestern.edu}\\
}

\begin{document}
\maketitle
\begin{abstract}


Active recognition, which allows intelligent agents to explore observations for better recognition performance, serves as a prerequisite for various embodied AI tasks, such as grasping, navigation and room arrangements.
Given the evolving environment and the multitude of object classes, it is impractical to include all possible classes during the training stage.
In this paper, we aim at advancing active open-vocabulary recognition, empowering embodied agents to actively perceive and classify arbitrary objects.
However, directly adopting recent open-vocabulary classification models, like Contrastive Language Image Pretraining (CLIP), poses its unique challenges.
Specifically, we observe that CLIP's performance is heavily affected by the viewpoint and occlusions, compromising its reliability in unconstrained embodied perception scenarios. 
Further, the sequential nature of observations in agent-environment interactions necessitates an effective method for integrating features that maintains discriminative strength for open-vocabulary classification.
To address these issues, we introduce a novel agent for active open-vocabulary recognition.
The proposed method leverages inter-frame and inter-concept similarities to navigate agent movements and to fuse features, without relying on class-specific knowledge.
Compared to baseline CLIP model with 29.6\% accuracy on ShapeNet dataset, the proposed agent could achieve 53.3\% accuracy for open-vocabulary recognition, without any fine-tuning to the equipped CLIP model.
Additional experiments conducted with the Habitat simulator further affirm the efficacy of our method.
\end{abstract}
\vspace{-4pt}

\section{Introduction}
\label{sec:intro}

\begin{figure}[t]
    \centering
    \begin{subfigure}{1\linewidth}
        \includegraphics[width=1\linewidth]{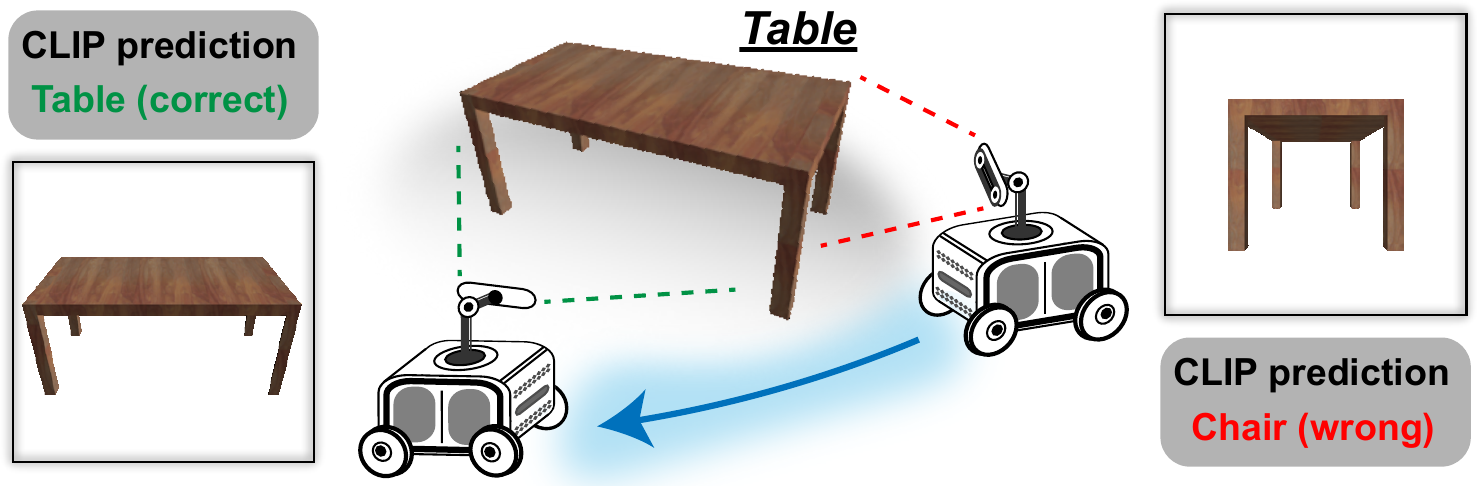}
        \caption{An illustrative episode of active recognition: An agent begins from a random viewpoint and has the freedom to execute movements to gather and aggregate information, thereby enhancing recognition performance. In this example, the integrated CLIP model fails to deliver an accurate prediction from the starting position, necessitating a change in viewpoint.}
        \label{fig:first_sight}
    \end{subfigure}
    \hfill
    \begin{subfigure}{1\linewidth}
        \includegraphics[width=1\linewidth]{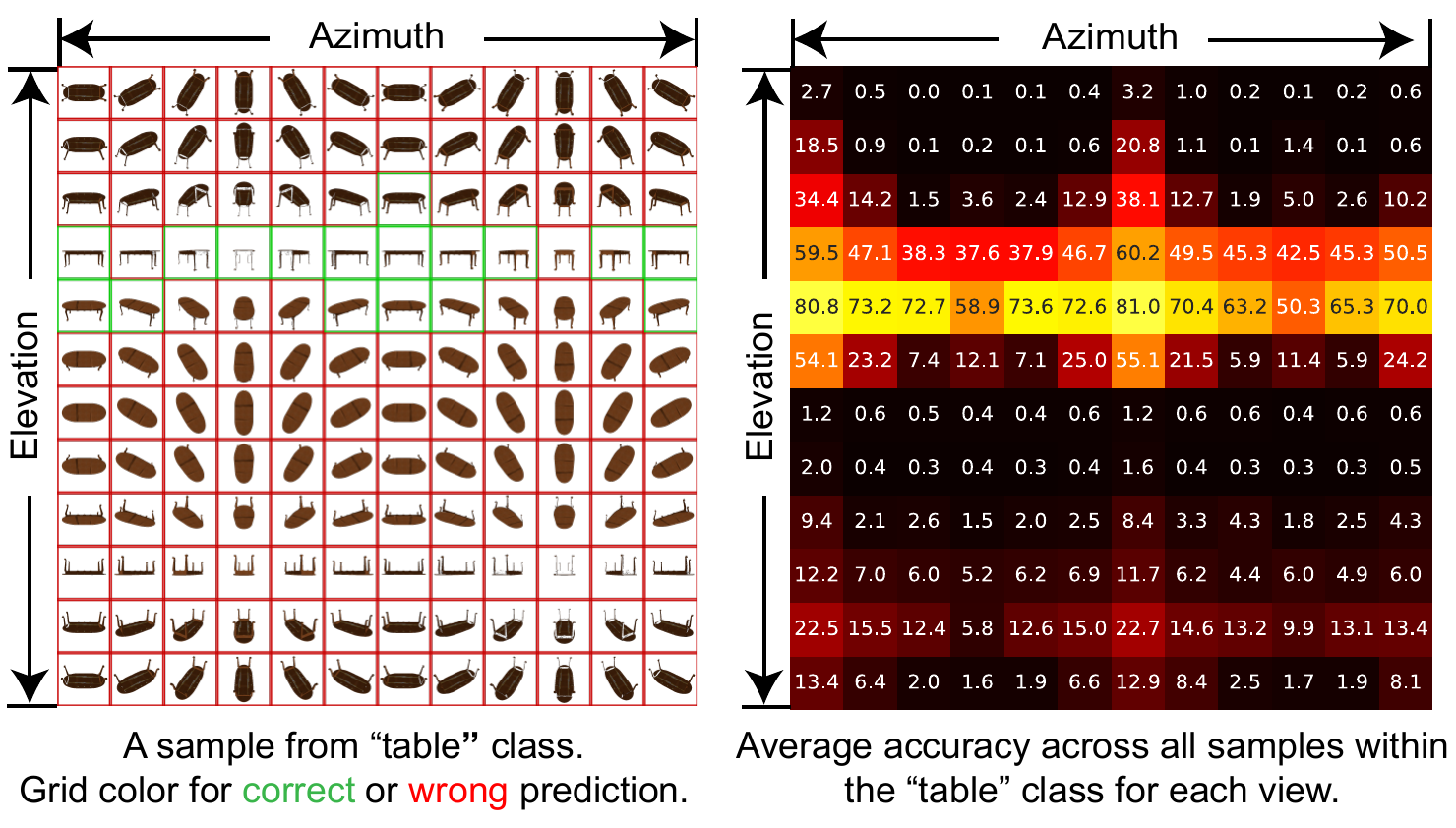}
        \caption{The performance of CLIP on the "table" class, collected from the ShapeNet dataset. We discretize the viewing sphere surrounding each object to a $12\times12$ viewing grid. The heatmap reveals a significant imbalance in accuracy across various viewpoints, underscoring the importance of active observation selection in embodied agents equipped with CLIP.}
        \label{fig:heatmap}
    \end{subfigure}
    \hfill
    \begin{subfigure}{1\linewidth}
        \includegraphics[width=1\linewidth]{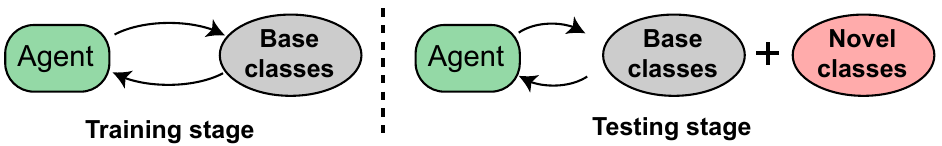}
        \caption{The task setting of active open-vocabulary recognition.}
        \label{fig:task_aovr}
    \end{subfigure}
    \caption{Illustrations of active open-vocabulary recognition task and the constraints of CLIP in embodied perception contexts.}
    \label{fig:clip_limit}
    \vspace{-8pt}
\end{figure}
In contrast to passive visual recognition~\cite{yue2015beyond,he2016deep,dosovitskiy2020image,liu2022video,ding2023learning}, where the input is obtained through human-operated or stationary devices, an active recognition agent is endowed with the ability to actively seek different observations, as illustrated in Figure~\ref{fig:first_sight}.
This dynamic approach effectively mitigates challenges associated with undesirable viewing conditions, including ambiguous viewpoints and occlusions, by employing a sequence of intelligent movements.


Recent advancements have spotlighted active recognition~\cite{jayaraman2016look,cheng2018geometry,yang2019embodied,fan2021flar} as a critical component in a range of embodied AI tasks~\cite{kolve2017ai2,szot2021habitat,majumdar2022zson,khandelwal2022simple}, extending from fundamental object grasping~\cite{chen2020transferable} to the more intricate task of semantic goal navigation~\cite{wortsman2019learning,chaplot2020object,pal2021learning,majumdar2022zson,zhao2023zero}. 
Consider the scenario of a semantic-goal navigation agent: it must actively recognize various objects while navigating towards its target. 
Although numerous studies have shown promising results in active recognition, these methods predominantly operate within a closed-vocabulary framework. 
This limitation means that the agent is restricted to recognizing only a pre-defined set of object classes. 
However, the real-world environments in which embodied agents operate, such as households, encompass a vast number of object classes.
This diversity renders it impractical to include all possible categories during the training phase.

Open-vocabulary recognition approaches~\cite{zareian2021open,gu2021open,minderer2022simple,zang2022open} are designed to overcome the limitations inherent in traditional models by aligning latent image-text embeddings. 
This allows for object classification using arbitrary text inputs during the testing phase. 
A notable advancement in this domain is the recent work on CLIP~\cite{radford2021learning,xu2023demystifying,schuhmann2022laion,cherti2023reproducible}, which elevates open-vocabulary recognition performance by collecting millions of image-text pairs and employing contrastive learning techniques. 
While CLIP represents a significant leap forward in single-image open-vocabulary recognition, its direct application as a visual recognizer in embodied agents presents notable challenges.

Specifically, a successful open-vocabulary recognition on agents entails three requirements:
(1) Our investigations into CLIP’s performance across widely-utilized platforms~\cite{chang2015shapenet,savva2019habitat,szot2021habitat} for embodied AI tasks reveal a noticeable sensitivity to varying viewpoints and occlusion levels. 
This could undermine its reliability in embodied perception, as illustrated in Figure~\ref{fig:heatmap}. 
Consequently, an agent must possess the capability to make strategic decisions in exploring informative observations based on its current status. 
This requirement fits into the realm of active recognition.
(2) As the agent interacts with its environment, it accumulates a sequence of observations. These observations require an effective integration mechanism to facilitate accurate class prediction, encompassing both base and novel classes.
(3) The intelligent perceiving policy guiding the agent’s recognition process should demonstrate robust generalization capabilities when encountering novel categories during testing.

In this paper, we introduce a novel approach for active open-vocabulary recognition, a critical yet under-explored area in embodied perception research. 
The task setting is depicted in Figure~\ref{fig:task_aovr}.
Our method leverages a reinforcement learning framework, enabling the agent to interact with its environment to learn the optimal recognition policy. 
Central to our approach is the idea of disentangling both policy and fusion method from class-specific representations. 
This strategy allows the model to adeptly handle novel classes encountered during testing.
We postulate that recognition policies should vary based on the semantic differences between classes. 
For instance, the policy for recognizing a cat and a dog are likely more aligned than those for a cat and a television. 
Based on this intuition, our policy input is grounded in the semantic proximity between visual embeddings and the text embeddings of base categories.
Moreover, we incorporate a self-attention module to allocate appropriate weights to visual features derived from a sequence of observations. 
These weighted features are subsequently integrated to formulate the final prediction. 
Our fusion method enhances performance in two significant ways: Firstly, it allows for the exclusion of inaccurate or uncertain predictions from single frames, thus preventing them from adversely affecting the integrated feature. 
Secondly, it could retain the discriminative efficacy towards novel classes.




The contributions of this paper can be summarized as follows: 
(1) We investigate the limitations of CLIP in terms of viewpoints and occlusion levels, uncovering a compelling reason to actively explore different views when employing CLIP in embodied agents. 
This leads to the proposition of a new task, termed active open-vocabulary recognition, aimed at enhancing embodied visual recognition.
(2) To address generalization challenges, the proposed agent leverages categorical concept similarities and frame-wise similarities to guide its evidence integration and also navigation.
(3) The efficacy of our agent is thoroughly evaluated across various dimensions on two widely-used platforms, namely ShapeNet~\cite{chang2015shapenet} and Habitat~\cite{szot2021habitat}. 
Our ablation studies, focusing on different fusion strategies and policy inputs, further underscore the superiority of our method.



\section{Related Work}
\label{sec:related}

\noindent{\bf{Active recognition.}}
Active vision, a long-standing task driven by the goal of enabling agents to intelligently acquire visual observations, has been explored through various approaches~\cite{gallos2019active,chaplot2020object,yang2019embodied,gadre2022continuous,nilsson2021embodied}.
A notable branch of this field is active recognition~\cite{aloimonos1988active,aloimonos1990purposive,andreopoulos2009theory,cheng2018geometry,johns2016pairwise} or detection~\cite{fang2020move,kotar2022interactron,ding2023learning}, a task that empowers an agent to gather observations driven by its own motives, thereby enhancing recognition performance.
Compared to the next-best-view problem~\cite{wu20143d,doumanoglou2016recovering}, active recognition aims for longer-term results, with its evaluation hinging on the overall movement cost.


Recent approaches in active recognition~\cite{wortsman2019learning,jayaraman2016look,fan2023avoiding,ding2023learning} typically characterize the recognition process as a Markov Decision Process, adopting reinforcement learning to derive optimal policies.
For example, in~\cite{jayaraman2016look,jayaraman2018end}, the authors develop an active recognition agent with an additional look-ahead module to anticipate future observations.
In parallel, a more holistic approach to object understanding, termed embodied amodal recognition, is proposed in~\cite{yang2019embodied}, offering not only categorization but also amodal bounding boxes and masks. 
Recently, a novel training strategy for active object detection, which combines online and offline data with a decision transformer, is introduced in \cite{ding2023learning}.

However, the predominant active recognition approaches are typically limited to predefined categories. 
This constraint means that an agent is only capable of recognizing specific classes post-deployment. 
In~\cite{fan2021flar}, the authors attempt to address this by integrating continual learning into active recognition, thereby accommodating incrementally introduced classes.
Despite its motivation, this approach still requires continuous training for each new class and is prone to catastrophic forgetting.
In contrast, we tackle active open-vocabulary recognition in this paper, which surpasses the traditional confines of fixed categories and obviates the need for additional training after deployment.


\noindent{\bf{CLIP model.}}
The pioneering CLIP~\cite{radford2021learning} and its subsequent variants~\cite{schuhmann2022laion,xu2023demystifying,cherti2023reproducible} have demonstrated notable efficacy in zero-shot image recognition across various benchmarks~\cite{russakovsky2015imagenet,soomro2012ucf101}.
These models operate on the principle of aligning visual and textual inputs within a shared embedding space, achieved through training with extensive image-text pair datasets. 
In the testing phase, the model calculates cosine similarity between encoded visual features and a set of text embeddings, leading to predictions based on the highest similarity. 
Recent works~\cite{khandelwal2022simple,shah2023lm,majumdar2022zson} have integrated CLIP models, or their derivative features, into embodied AI applications, yielding substantial improvements compared to backbones pretrained on ImageNet.

While there has been considerable exploration into both the capabilities and limitations of CLIP models from various perspectives~\cite{radford2021learning,shen2021much}, studies on their performance under varying viewpoints or levels of occlusion remains sparse.
Given that sub-optimal viewing conditions are common in embodied recognition scenarios, our study examines CLIP model performance under these conditions on two popular platforms~\cite{chang2015shapenet,szot2021habitat}. 
We find that the impact of adverse viewing conditions is significant. 
Consequently, for effective deployment of CLIP models as open-vocabulary recognizers on embodied agents, the ability for acquiring diverse observations, \ie, active recognition, is essential.


\noindent{\bf{Zero-shot object navigation.}} 
It is a task that aims to guide a robot to find a target belonging to an unseen class~\cite{majumdar2022zson,zhao2023zero}.
This task diverges from active recognition; it emphasizes guiding the robot to approach the target by leveraging prior semantic relations, such as the higher likelihood of finding a mug in a kitchen rather than in a bathroom.

To conclude this section, we highlight our dual motivations in active open-vocabulary recognition, which are intrinsically synergistic.
Our primary aim is to empower active recognition agents with the proficiency to effectively manage unseen classes, leveraging the capabilities of CLIP models.
Concurrently, we strive to address the limitations of CLIP, such as handling viewpoint variations and occlusions, by integrating active recognition strategies. 

\section{When is CLIP ineffective?}
\label{sec:limit}


Consider an embodied agent working in a household environment: the object of interest could be cluttered, creating heavy occlusions; located far from reach, resulting in an obscure sight; or relatively positioned in an ambiguous viewpoint.
These undesired viewing conditions are prevalent in embodied perception scenarios because we cannot control the environment to observe or the capturing setup of the embodied agent.
Similar challenges exist in related fields, such as first-person vision~\cite{dunnhofer2021first} and autonomous driving~\cite{zablocki2022explainability}.

Consequently, it is imperative for the recognition systems used by embodied agents to adeptly handle such challenges. 
Before deploying CLIP in these scenarios, it is crucial to thoroughly evaluate its performance under conditions of embodied perception. 
Our study focuses on two key aspects of adverse viewing conditions: varied viewpoints and the presence of occlusions. 

To conduct this evaluation, we select two CLIP models with Vision Transformers (ViT-B/32, ViT-L/14), one model based on ResNet-50 architecture (RN50x64)~\cite{radford2021learning}, and the recently developed MetaCLIP~\cite{xu2023demystifying} for detailed examination. 
These models will be assessed using the zero-shot recognition approach on specifically curated datasets.
Due to the constraint in page length, the main paper will primarily present the results pertaining to the ViT-B/32 model, while the analyses of the other model architectures will be detailed in our supplementary materials. 


\subsection{Datasets for investigation}
\label{sec:create_data}
We collect testing datasets from two widely-adopted platforms~\cite{chang2015shapenet,szot2021habitat} for testing varying viewpoints and occlusions.

\begin{figure}[t]
    \centering
    \begin{subfigure}[t]{0.40\linewidth}
        \includegraphics[width=1\linewidth]{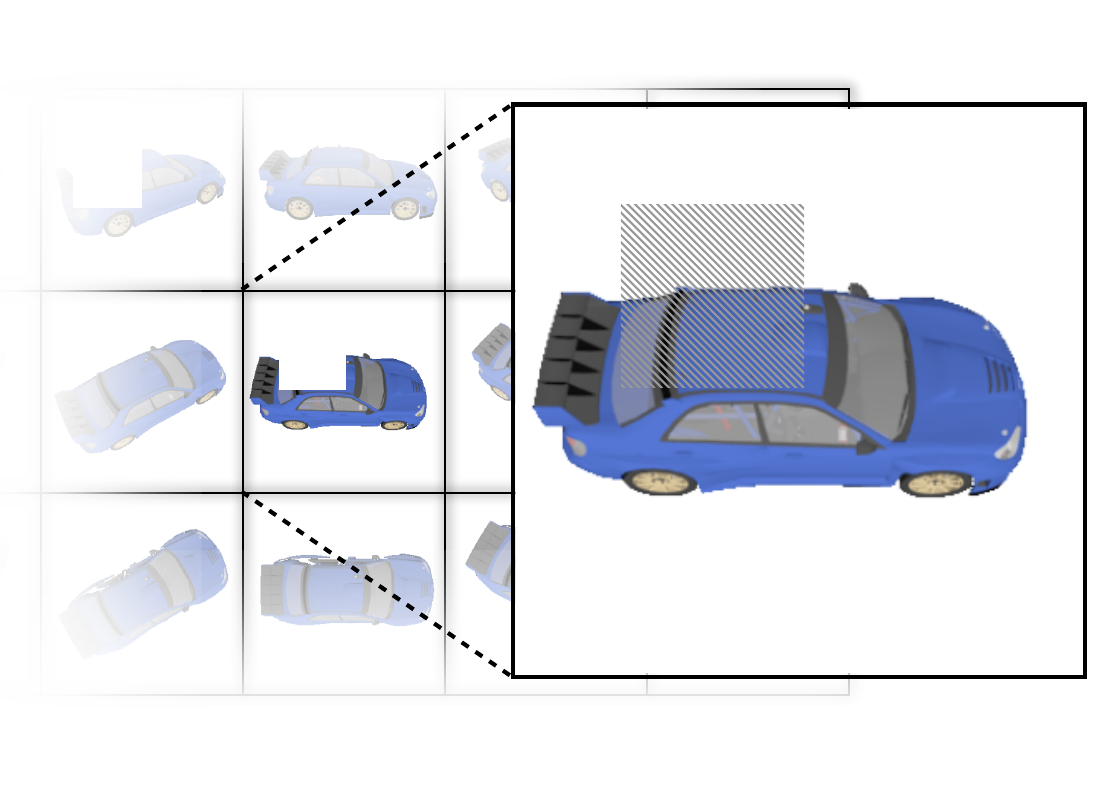}
        \caption{Occlusions by creating random masks for ShapeNet.}
        \label{fig:shapenet_occ}
    \end{subfigure}
    \hfill
    \begin{subfigure}[t]{0.58\linewidth}
        \includegraphics[width=1\linewidth]{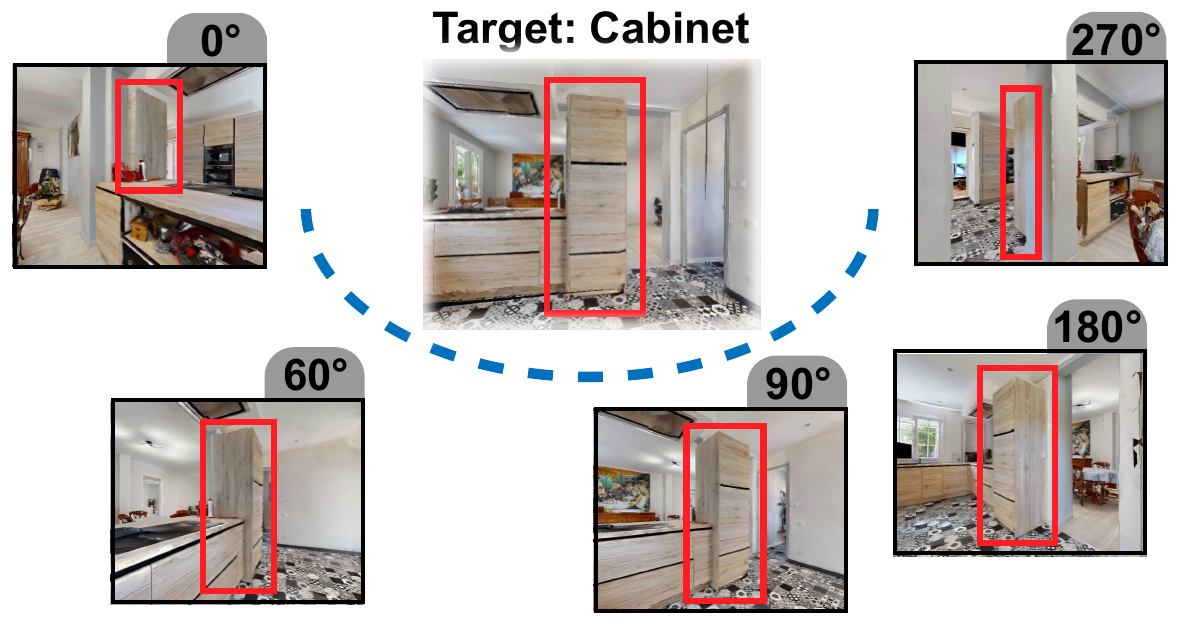}
        \caption{The collecting process of different viewpoints in Habitat.}
        \label{fig:habitat_collect}
    \end{subfigure}
    \caption{The process of curating test datasets to evaluate CLIP performance with occlusions and varying viewpoints.}
    \vspace{-8pt}
    \label{fig:curated_dataset}
\end{figure}

\begin{figure*}[t]
    \centering
    \includegraphics[width=1\linewidth]{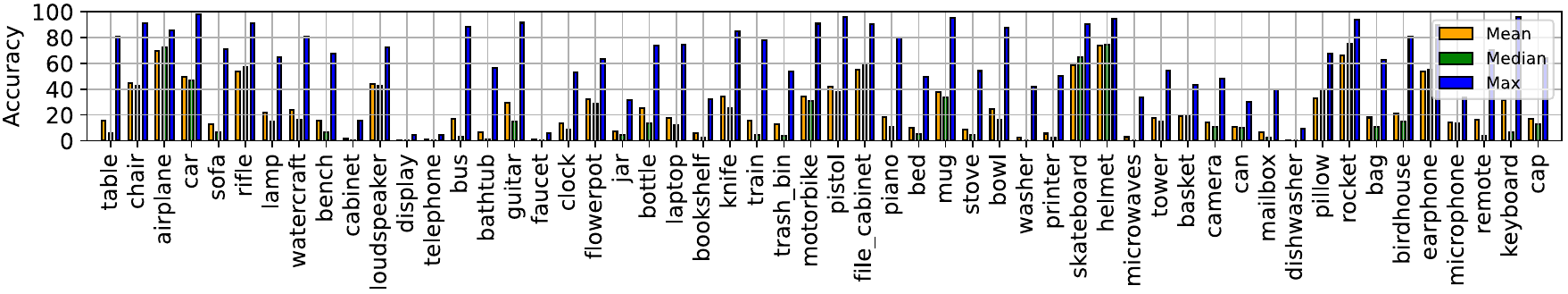}
    \caption{Performance of CLIP across all viewpoints within each category, reporting the mean, median, and maximum accuracy. The discrepancy between mean and maximum accuracy highlights CLIP's potential sensitivity to different viewpoints.}
    \vspace{-8pt}
    \label{fig:bars_sh}
\end{figure*}
\noindent{\bf{ShapeNet dataset.}}
The ShapeNetCore dataset~\cite{chang2015shapenet} comprises approximately $41500$ Computer-Aided Design (CAD) models spanning 55 common categories in their training split.
This dataset is utilized for studying active object recognition~\cite{jayaraman2016look,jayaraman2018end,fan2021flar}, as it allows an agent to manipulate 3D objects through movements, thereby acquiring novel observations.
To analyze the performance of CLIP across various viewpoints, we discretize the viewing sphere surrounding each object into $30$-degree segments. 
This approach results in a viewing grid with $M=12$ azimuths and $N=12$ elevations for each object. 
An example of viewing grid is shown on the left side of Figure~\ref{fig:heatmap}. 
Another rationale for selecting the ShapeNet dataset is its geometric alignment within classes. 
All 3D models in a given class are aligned, ensuring that identical grid coordinates correspond to comparable viewpoints across different samples.

To examine the impact of occlusions, we introduce random sight-blocking patches at each viewpoint, applying them with a predetermined probability. 
Each patch, defined as a $\frac{1}{3}$-length square of the original view, is randomly positioned within the current visual field. 
An example of a contaminated viewing grid is displayed in Figure~\ref{fig:shapenet_occ}. 
It is important to note that the level of occlusion intensifies as the probability of view-blocking increases.


\begin{figure}[t]
    \centering
    \includegraphics[width=1\linewidth]{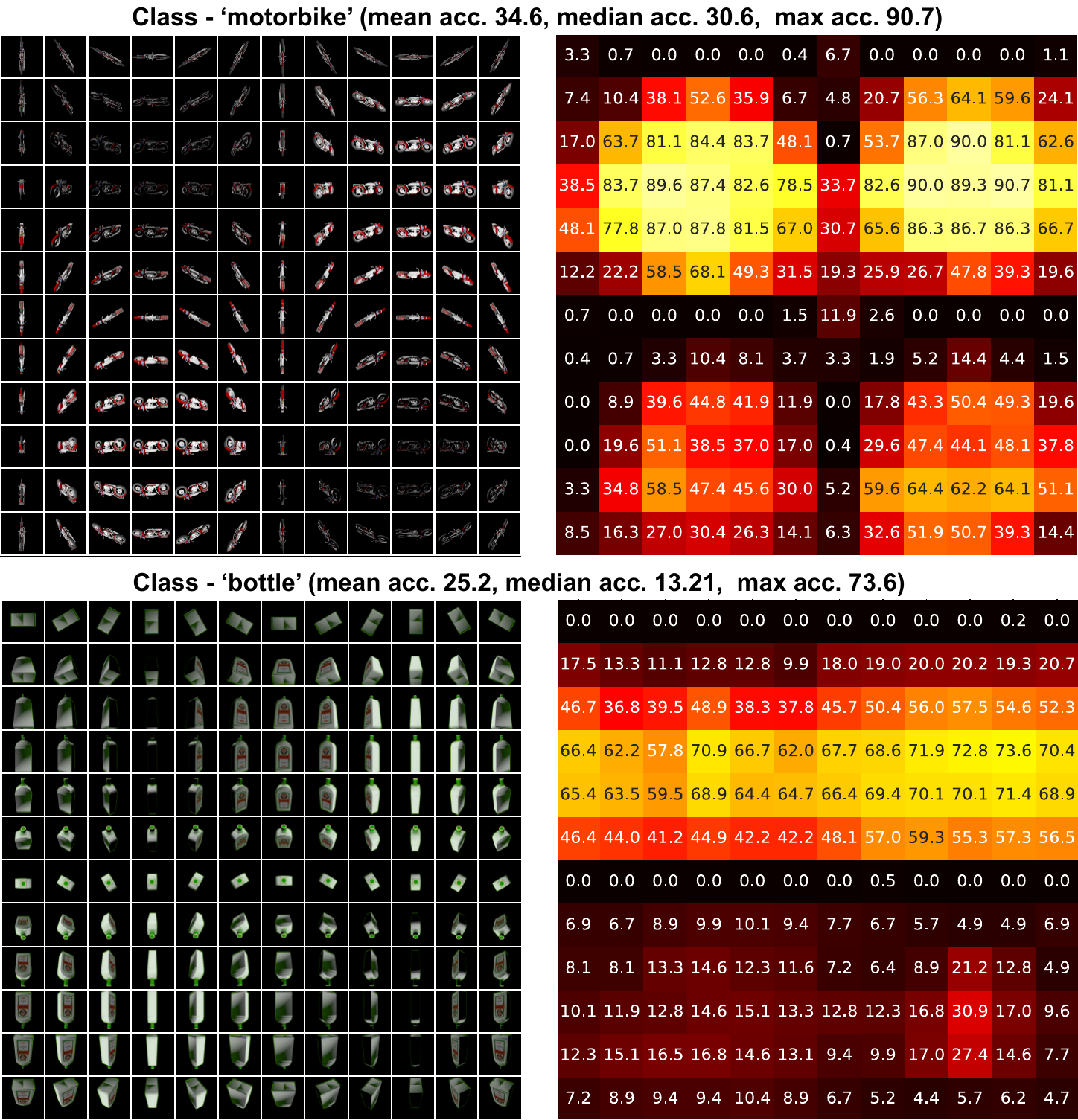}
    \caption{Recognition accuracy of different viewpoints on 'motorbike' and 'bottle' classes. The left side shows a testing sample for each class, while the right heatmap displays the average accuracy across all testing samples for each viewpoint.}
    \vspace{-8pt}
    \label{fig:more_heatmap}
\end{figure}

\noindent{\bf{Habitat dataset.}}
To examine the efficacy of CLIP in complex indoor environments, we construct a testing dataset using the existing simulator~\cite{szot2021habitat} with $145$ semantic annotated scenes. 
From $40$ labeled categories, we select $25$, excluding those that are ambiguous or related to building components such as doors, walls, and ceilings.
This approach results in a dataset comprising $4659$ objects. 
Detailed class distribution information is available in our supplementary materials.
For each chosen object, the agent is positioned at a random location, facing the target within a maximum distance of $3.0$ meters. 
Subsequently, the agent rotates in $30$-degree increments around the target on the horizontal plane. 
It is important to note that images where the target is not visible are excluded from the test. 
The image capture process is generally illustrated in Figure~\ref{fig:habitat_collect}.



\subsection{Sensitivity of CLIP to viewpoints}
We first examine the sensitivity of CLIP to viewpoints, utilizing the ShapeNet dataset. 
We use all $12\times12$ views of each sample to test the CLIP's recognition with text inputs describing all categories, akin to zero-shot recognition. 
Given that samples within the same category are aligned in 3D, we compute the average accuracy for each specific viewpoint across all samples of the same class. 
This process results in a detailed accuracy map for all viewpoints.
Examples are given in Figure~\ref{fig:more_heatmap}. 
Subsequently, we calculate the mean, median, and maximum accuracy across all viewpoints, with the results presented in Figure~\ref{fig:bars_sh}.


Ideally, a proper visual recognition system should exhibit robustness to changes in viewpoint, maintaining consistent performance across different angles. 
However, our analysis of CLIP's performance reveals significant variations in recognition accuracy depending on the viewpoint of the same object. 
Specifically, the average difference between the mean and maximum accuracy across all viewpoints and classes is a substantial $40.1\%$, a noteworthy level considering the overall average accuracy stood at $23.8\%$. 
Consequently, this suggests that an embodied agent using CLIP and adopting a random viewing position could experience a significant decline in performance compared to an optimal viewpoint.

\begin{figure}[t]
    \centering
    \includegraphics[width=1\linewidth]{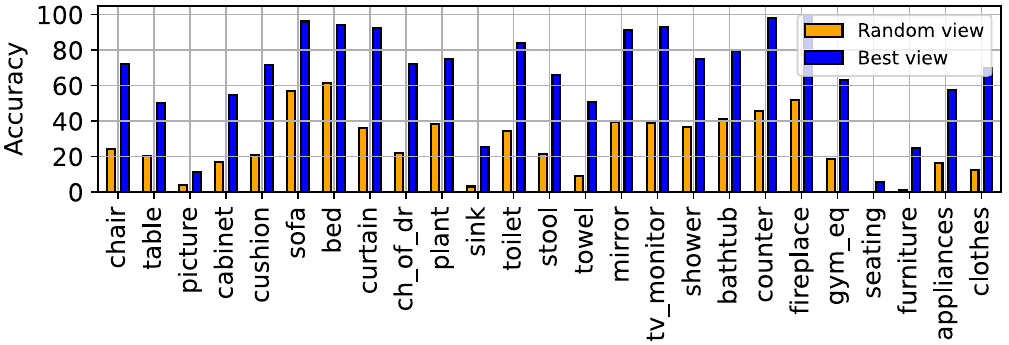}
    \caption{The accuracy comparison on the Habitat dataset between randomly taking viewpoints (average of 5 runs) and the best performing viewpoint.}
    \vspace{-8pt}
    \label{fig:bars_hb}
\end{figure}

Further exploration is conducted using the collected Habitat dataset, which features multi-view object images from an indoor simulator. 
Unlike the ShapeNet dataset, images in the Habitat dataset are captured at the horizontal plane and include natural occlusions typical of indoor settings. 
Since objects within the same class in the Habitat dataset are not aligned, we assessed the accuracy gap on a per-sample basis. 
For each object, we record the highest performance across all available viewpoints and compare it with the accuracy of a randomly selected view. 
The findings, depicted in Figure~\ref{fig:bars_hb}, show an average discrepancy of $40.2\%$ between the performances of the random and optimal views, with an average random view accuracy of $26.8\%$ across classes.



\subsection{Impact of occlusions on CLIP}
\begin{figure}[t]
    \centering
    \includegraphics[width=1\linewidth]{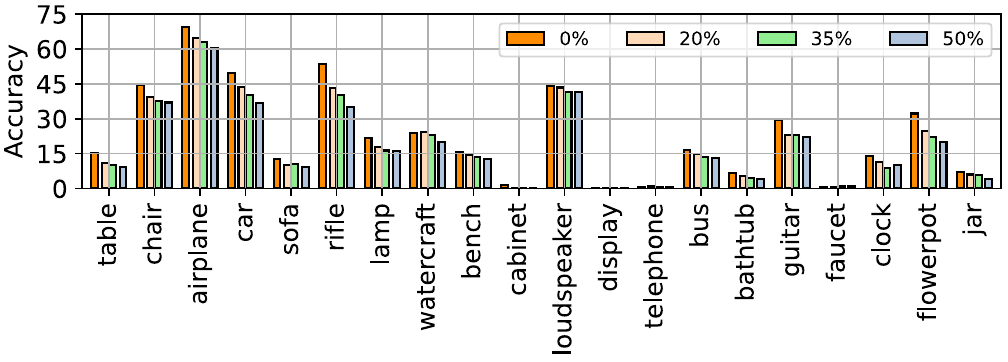}
    \caption{The mean accuracy decline observed when various occlusion levels ($20\%$, $35\%$, $50\%$) are applied to the ShapeNet dataset.}
    \vspace{-8pt}
    \label{fig:limit_occ}
\end{figure}

We conducted a preliminary experiment to assess the impact of occlusions on the performance of CLIP. 
Three distinct levels of random occlusions are applied to the ShapeNet dataset, as outlined in Section~\ref{sec:create_data}. 
Figure~\ref{fig:limit_occ} illustrates the decrease in mean accuracy across the most common $20$ classes. 
Notably, the average accuracy drop across all $55$ classes at three different occlusion levels are $3.1\%$, $4.0\%$, and $5.0\%$, respectively. 
Although the occlusions introduced in this experiment are relatively simple compared to the more challenging occlusions encountered in embodied recognition scenarios, they nonetheless underscore the necessity of intelligent perceiving for enhancing CLIP's robustness to occlusions.

\section{Active Open-Vocabulary Recognition}
\label{sec:method}
\begin{figure*}[t]
    \centering
    \includegraphics[width=0.92\linewidth]{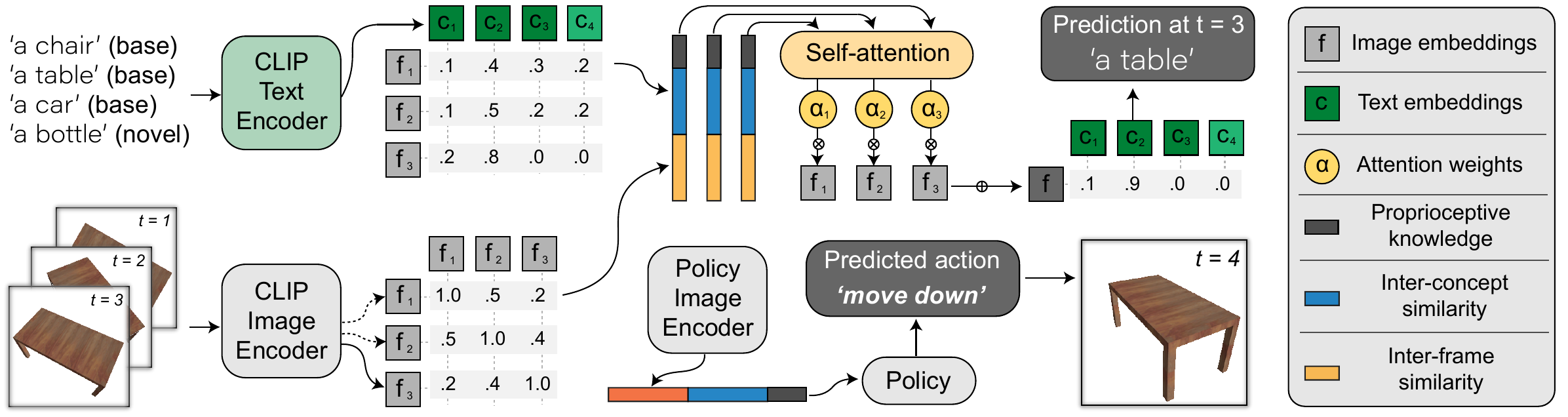}
    \caption{An illustration of the proposed architecture for the active open-vocabulary recognition agent. Only key data flows are depicted to avoid confusion. The meanings of different features and weights are detailed in the grey box on the right.}
    \vspace{-8pt}
    \label{fig:method}
\end{figure*}

In this section, we introduce our proposed agent, designed to handle active open-vocabulary recognition. 
We outline three essential requirements for an agent to effectively manage this challenging task. 
Firstly, the agent must intelligently perceive novel and informative views to enhance recognition performance as discussed in Section~\ref{sec:limit}.
Secondly, it is important to effectively integrate accumulated evidence from observations, including for novel categories not encountered during the training phase. 
Thirdly, the recognition policy should also demonstrate a robust generalization capability towards novel categories.
To meet these requirements, our method models the recognition process within the framework of reinforcement learning. 
Here, the policy is rewarded for successful recognition.
Moreover, rather than recurrently aggregating single-image features for recognition~\cite{jayaraman2016look,fan2021flar}, our approach involves assigning weights to CLIP features using a self-attention module. 
This strategy helps to preserve the inherent capability of recognizing novel categories without adverse interference.
For the policy, we incorporate a similarity measure that compares the current image with text embeddings from base classes, thereby highlighting their semantic differences.
An illustrative overview of our proposed agent can be found in Figure~\ref{fig:method}.

\subsection{Problem setup and notation}

We outline our problem setup by initially introducing a basic active recognition agent, subsequently extending this concept to the open-vocabulary recognition scenario.

The active recognition agent is provided with a target object, denoted as $x$, associated with an unknown label $y$. 
The agent is permitted a total of $T$ steps to make the final prediction $\hat{y}$ of the target. 
At each timestep $t = 1, \dots, T-1$, the agent may execute an additional action $a^t \in \mathcal{A}$, signifying a movement to alter its viewing point for a new observation $v_t$. 
Here, $\mathcal{A}$ represents the predefined action space, \eg, elevating the camera by $30$ degrees. 
Through these movements, the agent aims to enhance its recognition performance by efficiently exploring its environment, aggregating observations, and classifying based on the integrated information.

Following the introduction of the active recognition task, we discuss the open-vocabulary recognition setup. 
During the training phase, the object presented to the agent is sampled from a set of base classes, $\mathcal{C}_B$. 
Conversely, in the testing phase, the target object is sourced from a broader open vocabulary, $\mathcal{C}_O$, where $\mathcal{C}_B \subset \mathcal{C}_O$. 
The novel classes, not included in the base classes, are collectively referred to as $\mathcal{C}_N$. 
Additionally, for each class within $\mathcal{C}_O$, a corresponding text embedding is available during testing, as current vision-language models, such as CLIP, necessitate a similarity measure to predict the class.


\subsection{Attention-based feature integration}

The image CLIP model, as described in~\cite{radford2021learning}, lacks the capability to aggregate temporal information while producing per-step features. 
For a recognition episode of the target $x$, we obtain a set of observations denoted as $\mathcal{V}_x=\{v_1,\dots,v_T\}$.
We omit the subscript $x$ in subsequent formulations for clarity.
The corresponding image embeddings are denoted as a sequence $\mathcal{F}=(f_1,\dots,f_T)$, obtained from a fixed CLIP image encoder.

Each feature $f$ could be directly utilized to predict the class of the target.
However, to counter undesired viewing conditions and enhance robustness, it is critical to integrate evidence across frames.
We employ an attention mechanism to derive the global feature, which can be formulated as:
\begin{equation}
    f=\mathcal{F}\alpha,
\end{equation}
where $\alpha$ represents the attention weight vector of dimension $T$, satisfying $\ell_1{\text{-norm}}=1$.
The weight $\alpha$ assesses the importance of each frame, filtering irrelevant information and retaining critical details in the global feature.

To obtain $\alpha$, we utilize a self-attention module and a linear layer.
These map input features $(q_1,\dots, q_T)$ to a $T$-dimensional embedding, followed by the softmax function to finalize $\alpha$.
The design of the input feature $q_t$ is crucial, as it must avoid introducing class-specific knowledge into the self-attention module.
Therefore, we use the top-$k$ cosine similarity between the image feature $f_t$ and text embeddings $\{c_i,i\in\mathcal{C}_O\}$, forming the inter-concept similarity $s^{\text{concept}}_t$.
This similarity, devoid of specific semantics, maintains a measure of recognition confidence for the current frame.
Specifically, a skewed distribution of inter-concept similarity usually indicates the feature $f_t$ contains low ambiguities.
It is also important to mention that the inter-concept similarity is a preliminary approach for estimating uncertainties in CLIP features, which presumes that CLIP rarely make overconfident mis-classifications.

In addition to $s^{\text{concept}}_t$, we incorporate the inter-frame similarity $s^{\text{frame}}_t$ and proprioceptive knowledge $p_t$ into the self-attention module.
The inter-frame similarity $s^{\text{frame}}_t$ quantifies the semantic distances between the current feature $f_t$ with features from other frames.
Meanwhile, $p_t$ details the relative location change compared to the previous frame and the last action $a_{t-1}$ taken.
Consequently, our input feature for the self-attention module is $q_t=[s^{\text{concept}}_t, s^{\text{frame}}_t, p_t]$.
During training with base classes $\mathcal{C}_B$, we apply the following loss to the global feature $f$:
\begin{equation}
    \mathcal{L}_{\text{attention}}=F_{\text{cross-entropy}}(\hat{y},y),
\end{equation}
where $\hat{y}$ is the prediction obtained by applying the softmax function on $\{fc_i^{\intercal},i\in\mathcal{C}_B\}$.

\subsection{Active open-vocabulary recognition agent}

We introduce the approach enabling the agent to intelligently navigate in order to acquire informative observations, namely, the policy module.
The policy is formulated as a Partially Observable Markov Decision Process (POMDP), represented by the pdf $\pi(a_t | h_{t-1})$. 
Here, $h_{t-1}$ represents the agent's prior state, recurrently derived from observations up to time step $t-1$. 
Specifically, the policy module integrates a single-layer Gated Recurrent Unit (GRU) to accumulate temporal information, and two linear layers function as actor and critic based on the GRU output. 
For the policy input, we employ a separate image encoder, \ie, a simple 3-layer network, to generate the image embedding $f^{\text{policy}}_t$. 
This choice, instead of using CLIP features, is to preclude the infusion of strong semantic knowledge into the policy, which could potentially impede its generalization. 
Analogous to our feature integration module, the policy input is incorporated with inter-concept similarity $s^{\text{concept'}}_t$ and proprioceptive knowledge $p_t$. 
However, the $s^{\text{concept'}}_t$ for the policy solely focuses on the base classes $\mathcal{C}_B$, excluding the top-$k$ operation, thus yielding a $|\mathcal{C}_B|$-dimensional embedding that reflects the current observation's relation to base classes.


The training reward for the policy is defined as the classification score of the ground-truth class $y$, which varies between $0$ to $1$.
This reward changes based on the precision of predictions for the correct classes, which is derived from the global feature. 
We adopt the Proximal Policy Optimization (PPO) algorithm~\cite{schulman2017proximal} to train the recognition policy from the interactions between the agent and its environment.


\begin{table*}[t]
\centering
\scriptsize
\caption{Recognition success rates on the ShapeNet dataset with varied class splits. 
The success rate is measured based on the final predictions. 
The difference between our method and other heuristic policies, \ie, {\texttt{Random}}, {\texttt{Largest-step}}, highlights the improvements achieved through intelligent movements. 
{\texttt{Last-prediction}} represents the last-step prediction of our agent without evidence fusion.}
\label{tab:main}
\begin{tabular}{c|cccccccccccccc}
\hline
\multirow{4}{*}{\textbf{Model}} &
  \multicolumn{14}{c}{\textbf{Base/novel/open classes split}} \\ \cline{2-15} 
 &
  \multicolumn{6}{c|}{\textbf{10/45/55}} &
  \multicolumn{6}{c|}{\textbf{20/35/55}} &
  \multicolumn{2}{c}{\textbf{55/0/55}} \\ \cline{2-15} 
 &
  \multicolumn{2}{c|}{\textbf{Base classes}} &
  \multicolumn{2}{c|}{\textbf{Novel classes}} &
  \multicolumn{2}{c|}{\textbf{Open classes}} &
  \multicolumn{2}{c|}{\textbf{Base classes}} &
  \multicolumn{2}{c|}{\textbf{Novel classes}} &
  \multicolumn{2}{c|}{\textbf{Open classes}} &
  \multicolumn{2}{c}{\textbf{Base classes}} \\ \cline{2-15} 
 &
  \multicolumn{1}{c|}{\textit{top-1}} &
  \multicolumn{1}{c|}{\textit{top-3}} &
  \multicolumn{1}{c|}{\textit{top-1}} &
  \multicolumn{1}{c|}{\textit{top-3}} &
  \multicolumn{1}{c|}{\textit{top-1}} &
  \multicolumn{1}{c|}{\textit{top-3}} &
  \multicolumn{1}{c|}{\textit{top-1}} &
  \multicolumn{1}{c|}{\textit{top-3}} &
  \multicolumn{1}{c|}{\textit{top-1}} &
  \multicolumn{1}{c|}{\textit{top-3}} &
  \multicolumn{1}{c|}{\textit{top-1}} &
  \multicolumn{1}{c|}{\textit{top-3}} &
  \multicolumn{1}{c|}{\textit{top-1}} &
  \textit{top-3} \\ \hline\hline
CLIP (ViT-B/32)~\cite{radford2021learning} &
  \multicolumn{1}{c|}{33.1} &
  \multicolumn{1}{c|}{52.2} &
  \multicolumn{1}{c|}{21.6} &
  \multicolumn{1}{c|}{34.0} &
  \multicolumn{1}{c|}{29.6} &
  \multicolumn{1}{c|}{46.7} &
  \multicolumn{1}{c|}{30.1} &
  \multicolumn{1}{c|}{47.4} &
  \multicolumn{1}{c|}{24.8} &
  \multicolumn{1}{c|}{39.3} &
  \multicolumn{1}{c|}{29.6} &
  \multicolumn{1}{c|}{46.7} &
  \multicolumn{1}{c|}{29.6} &
46.7 \\ \hline
Ours + {\texttt{Random}} &
  \multicolumn{1}{c|}{39.3} &
  \multicolumn{1}{c|}{63.1} &
  \multicolumn{1}{c|}{26.5} &
  \multicolumn{1}{c|}{42.5} &
  \multicolumn{1}{c|}{35.4} &
  \multicolumn{1}{c|}{56.9} &
  \multicolumn{1}{c|}{35.4} &
  \multicolumn{1}{c|}{55.8} &
  \multicolumn{1}{c|}{29.0} &
  \multicolumn{1}{c|}{48.9} &
  \multicolumn{1}{c|}{35.4} &
  \multicolumn{1}{c|}{56.9} &
  \multicolumn{1}{c|}{35.4} & 56.9
   \\ \hline
Ours + {\texttt{Largest-step}} &
  \multicolumn{1}{c|}{41.0} &
  \multicolumn{1}{c|}{65.6} &
  \multicolumn{1}{c|}{26.8} &
  \multicolumn{1}{c|}{42.5} &
  \multicolumn{1}{c|}{36.7} &
  \multicolumn{1}{c|}{58.7} &
  \multicolumn{1}{c|}{37.8} &
  \multicolumn{1}{c|}{58.2} &
  \multicolumn{1}{c|}{30.1} &
  \multicolumn{1}{c|}{48.2} &
  \multicolumn{1}{c|}{36.7} &
  \multicolumn{1}{c|}{58.7} &
  \multicolumn{1}{c|}{36.7} & 58.7
   \\ \hline
Ours + {\texttt{Last-prediction}} &
  \multicolumn{1}{c|}{54.1} &
  \multicolumn{1}{c|}{72.9} &
  \multicolumn{1}{c|}{32.0} &
  \multicolumn{1}{c|}{48.4} &
  \multicolumn{1}{c|}{47.4} &
  \multicolumn{1}{c|}{65.5} &
  \multicolumn{1}{c|}{55.2} &
  \multicolumn{1}{c|}{73.1} &
  \multicolumn{1}{c|}{43.4} &
  \multicolumn{1}{c|}{62.9} &
  \multicolumn{1}{c|}{53.6} &
  \multicolumn{1}{c|}{71.7} &
  \multicolumn{1}{c|}{57.0} & 74.8
  \\ \hline
Ours &
  \multicolumn{1}{c|}{\textbf{60.6}} &
  \multicolumn{1}{c|}{\textbf{81.3}} &
  \multicolumn{1}{c|}{\textbf{36.6}} &
  \multicolumn{1}{c|}{\textbf{55.1}} &
  \multicolumn{1}{c|}{\textbf{53.3}} &
  \multicolumn{1}{c|}{\textbf{73.4}} &
  \multicolumn{1}{c|}{\textbf{57.9}} &
  \multicolumn{1}{c|}{\textbf{76.8}} &
  \multicolumn{1}{c|}{\textbf{47.8}} &
  \multicolumn{1}{c|}{\textbf{69.0}} &
  \multicolumn{1}{c|}{\textbf{56.6}} &
  \multicolumn{1}{c|}{\textbf{75.7}} &
  \multicolumn{1}{c|}{\textbf{59.2}} & \textbf{78.8}
   \\ \hline
\end{tabular}
\end{table*}

\section{Experiment}
\label{sec:experiment}
The objectives of our experimental analysis are threefold:
(1) Assessing the effectiveness of our agent in managing the open-vocabulary recognition task.
(2) Evaluating the advantages of the intelligent policy in comparison to the static, single-frame CLIP model.
(3) Conducting ablation studies to analyze the impact of two key components, \ie, the feature integration method and the policy input.

\subsection{Datasets and experimental setup}

\noindent{\textbf{ShapeNet.}} We utilized the ShapeNet~\cite{chang2015shapenet} to train the agent, initializing the viewing grid identically to the introduced investigation dataset in Section~\ref{sec:create_data}. 
Starting from a random viewpoint, the agent is permitted to take a total of $T=6$ steps. 
At each step, the agent can navigate within a $5 \times 5$ grid relative to its current position.

For the open-vocabulary setting, the most common $10$ or $20$ object categories are chosen as the base classes. 
The training set, comprising samples from these base classes, is available to the agent during its training stage. 
For evaluation, all testing samples from $55$ classes are used.

\noindent{\textbf{Habitat.}} The training of agent is conducted using $145$ semantically-annotated indoor scenes from HM3D~\cite{szot2021habitat}, while testing is carried out on $90$ non-overlapping scenes from MP3D~\cite{chang2017matterport3d}, both datasets being part of the Habitat platform. 
In each training or testing episode, the agent is randomly placed in a scene with the target identified by a bounding box.
The agent takes movement to view the target and predict its class label.
The defined action space includes $\{{\texttt{move\_forward}}, {\texttt{turn\_left}}, {\texttt{turn\_right}}, {\texttt{look\_up}},\\{\texttt{look\_down}}\}$. 
Actions allow the agent to move $0.25$m, turn by $10$ degrees, or tilt by $10$ degrees. A total of $T=10$ steps is permitted for experiments on the Habitat platform.

We categorized the $25$ classes into $10$ base classes and $15$ novel classes, based on their frequency in the HM3D dataset.

\subsection{Result on ShapeNet}

\begin{figure*}[t]
    \centering
    \includegraphics[width=1.0\linewidth]{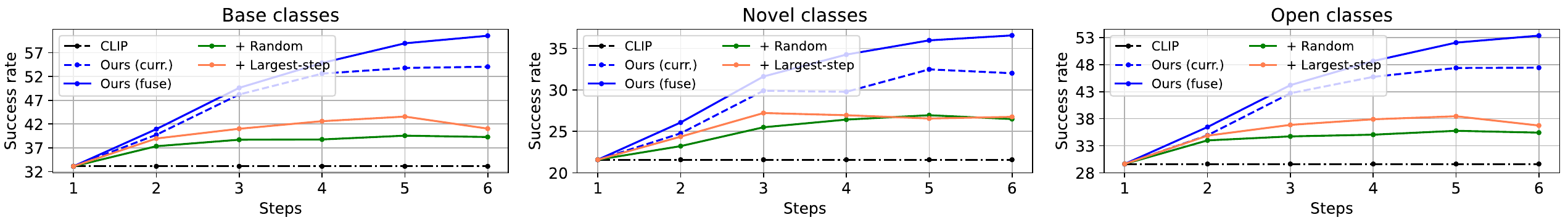}
    \caption{Performance comparison of different agents over steps on the ShapeNet dataset with the base/novel/open split of 10/45/55.}
    \vspace{-5pt}
    \label{fig:trend_shapenet}
\end{figure*}


We present the comparison of our proposed agent against other baselines in Table~\ref{tab:main}. 
In addition to CLIP, which serves as a passive recognition baseline, we utilize two heuristic policies, namely, \texttt{Random} and \texttt{Largest-step}, to highlight the benefits of integrating intelligent movements into the embodied recognition process. 
The \texttt{Random} policy randomly selects an action at each step, while \texttt{Largest-step} opts for a movement most distant from the current viewpoint. 
\texttt{Last-prediction}, based on our agent's last observation, indicates the success of our proposed policy in locating an informative view at its final step. 
Except for policy alterations, all compared baselines utilize the same architecture.

We report results for two different splits between base and novel classes for open-vocabulary recognition. Specifically, either $10$ or $20$ base classes are introduced during training. 
Test samples from all $55$ classes are used for evaluation. 
Furthermore, we present results when samples from all classes are available during training, serving as the upper bound.

Our agent achieves a $53.3\%$ success rate when trained with only $10$ classes, marking a $23.7\%$ absolute improvement over CLIP. 
Additionally, the success rate rises by $15.0\%$ when evaluating on the $45$ novel classes, underscoring our agent's effectiveness in handling unseen categories.
Our performance also surpasses other heuristic policies, demonstrating that random sequences of observations do not necessarily enhance recognition performance; instead, an intelligent policy is essential for effective embodied recognition.

Another finding relates to the variation in open-class results based on the number of base classes. 
Our method effectively learns an appropriate recognition policy even with limited base classes. 
For instance, when tested on open classes, our agent trained with only $10$ base classes experiences just a $3.3\%$ drop in success rate compared to training with $20$ classes. 
This resilience likely originates from our policy and evidence integration design, which avoids class-specific representations and instead relies on similarity measures.

The step-by-step performance with $10$ base classes is illustrated in Figure~\ref{fig:trend_shapenet}. In addition to the aforementioned baselines, we introduce another variant of our agent that predicts based solely on the current CLIP feature.
Its comparison with predictions using fused global features highlights the advantages of our attention-based integration method.

\subsection{Result on Habitat}
\begin{figure}[t]
    \centering
    \includegraphics[width=1\linewidth]{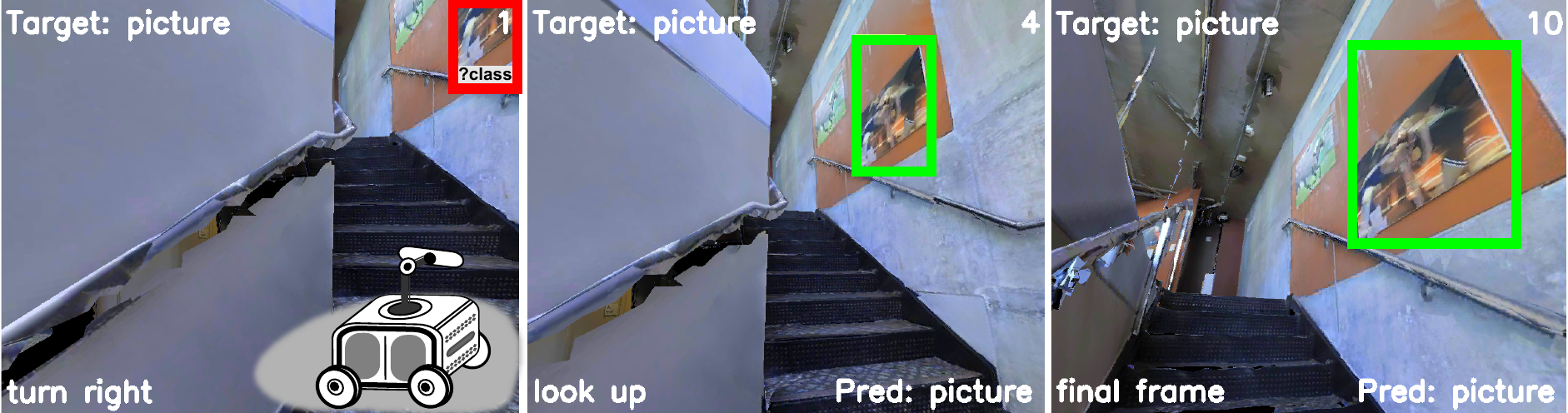}
    \caption{An episode of active recognition in the Habitat simulator. A target is queried at step $t=1$. The agent is permitted to make $10$ observations in each episode. Three steps (at $t = 1, 4, 10$) are shown, along with their subsequent actions and class predictions.}
    \vspace{-5pt}
    \label{fig:habitat_eps}
\end{figure}
Our evaluation on the Habitat simulator~\cite{szot2021habitat} involves allowing the agent to navigate freely within indoor environments to predict the class of a specified target. 
An testing episode is depicted in Figure~\ref{fig:habitat_eps}. 
In this study, we assess the performance of our recognition agent in comparison with CLIP and other heuristic-based policies. 
Additionally, we introduce a new policy, \texttt{Fixation}, which is designed to maintain the target at the center of view during movement. 
The results of these comparisons, presented in Table~\ref{tab:habitat}, further validates the effectiveness of our proposed method in complex scenarios.


\begin{table}[]
\centering
\scriptsize
\caption{Comparison of performance on the Habitat dataset using a class split of $10$ base classes and $15$ novel classes.}
\label{tab:habitat}
\begin{tabular}{c|cc|cc|cc}
\hline
\multirow{2}{*}{\textbf{Method}} & \multicolumn{2}{c|}{\textbf{Base classes}} & \multicolumn{2}{c|}{\textbf{Novel classes}} & \multicolumn{2}{c}{\textbf{Open classes}} \\ \cline{2-7} 
 & \multicolumn{1}{c|}{\textit{top-1}} & \textit{top-3} & \multicolumn{1}{c|}{\textit{top-1}} & \textit{top-3} & \multicolumn{1}{c|}{\textit{top-1}} & \textit{top-3} \\ \hline\hline
CLIP (ViT-B/32) & \multicolumn{1}{c|}{22.2} & 43.1 & \multicolumn{1}{c|}{32.3} & 55.0 & \multicolumn{1}{c|}{24.9} & 46.2 \\ \hline
Ours + \texttt{Random} & \multicolumn{1}{c|}{22.3} & 43.3 & \multicolumn{1}{c|}{32.5} & 55.2 & \multicolumn{1}{c|}{25.1} & 46.4 \\ \hline
Ours + \texttt{Fixation} & \multicolumn{1}{c|}{23.4} & 45.1 & \multicolumn{1}{c|}{33.7} & 56.3 & \multicolumn{1}{c|}{26.2} & 47.7 \\ \hline
Ours & \multicolumn{1}{c|}{\textbf{25.8}} & \textbf{48.8} & \multicolumn{1}{c|}{\textbf{35.4}} & \textbf{58.1} & \multicolumn{1}{c|}{\textbf{28.0}} & \textbf{49.8} \\ \hline
\end{tabular}
\vspace{-8pt}
\end{table}

\subsection{Ablation studies}

In this study, we investigate the effects of various factors, such as different evidence integration strategies and the choice of input features for the policy. 
The experiments are conducted using the ShapeNet dataset with $10$ base classes.


We assess four alternative evidence integration strategies in comparison with our proposed attention-based method: \textit{Average-feature}, \textit{Average-prediction}, \textit{Max-prediction}, and \textit{Vote}. 
The \textit{Average-feature} approach computes the mean of accumulated CLIP features across frames prior to making a prediction. 
In contrast, \textit{Average-prediction} averages the predictions from individual frames. 
The \textit{Max-prediction} method selects the single-frame prediction with the highest confidence as the final decision. 
And the \textit{Vote} technique employs a voting mechanism across all frame predictions to determine the final result. 
Table~\ref{tab:ablation} presents the results of these integration strategies. 
Our proposed method outperforms the alternatives, as the attention mechanism assigns varying weights to each frame, effectively reducing ambiguity in predictions while retaining critical information.


Moreover, we modify the input to our policy module during training to CLIP features, emphasizing the significance of disentangling semantic features for active open-vocabulary recognition. 
This alteration, however, adversely affects performance across all class splits, particularly in novel classes. 
The underlying reasons are twofold. 
First, while CLIP features offer a robust representation for classification, they may not be ideally suited for determining subsequent movements. 
Second, a policy input with strong semantic knowledge can potentially hinder the agent's performance in unfamiliar categories during testing.



\begin{table}[t]
\centering
\scriptsize
\caption{Ablation studies of evidence integration strategies and the policy input, evaluating their success rates at the final step.}
\label{tab:ablation}
\begin{tabular}{c|cc|cc|cc}
\hline
\multirow{2}{*}{\textbf{Method}} & \multicolumn{2}{c|}{\textbf{Base classes}} & \multicolumn{2}{c|}{\textbf{Novel classes}} & \multicolumn{2}{c}{\textbf{Open classes}} \\ \cline{2-7} 
 & \multicolumn{1}{c|}{\textit{top-1}} & \textit{top-3} & \multicolumn{1}{c|}{\textit{top-1}} & \textit{top-3} & \multicolumn{1}{c|}{\textit{top-1}} & \textit{top-3} \\ \hline\hline
Ours & \multicolumn{1}{c|}{\textbf{60.6}} & \textbf{81.3} & \multicolumn{1}{c|}{\textbf{36.6}} & \textbf{55.1} & \multicolumn{1}{c|}{\textbf{53.3}} & \textbf{73.4} \\ \hline
w/ \textit{Average-feature} & \multicolumn{1}{c|}{57.5} & 78.1 & \multicolumn{1}{c|}{34.7} & 54.0 & \multicolumn{1}{c|}{50.6} & 70.9 \\ \hline
w/ \textit{Average-prediction} & \multicolumn{1}{c|}{58.9} & 80.0 & \multicolumn{1}{c|}{35.9} & 54.3 & \multicolumn{1}{c|}{51.9} & 72.2 \\ \hline
w/ \textit{Max-prediction} & \multicolumn{1}{c|}{58.6} & 79.9 & \multicolumn{1}{c|}{35.2} & 53.7 & \multicolumn{1}{c|}{51.6} & 72.0 \\ \hline
w/ \textit{Vote} & \multicolumn{1}{c|}{58.8} & 80.1 & \multicolumn{1}{c|}{35.3} & 55.2 & \multicolumn{1}{c|}{51.7} & 72.8 \\ \hline\hline
Train w/ CLIP feature & \multicolumn{1}{c|}{52.6} & 74.7 & \multicolumn{1}{c|}{28.6} & 46.0 & \multicolumn{1}{c|}{45.4} & 66.1 \\ \hline
\end{tabular}
\vspace{-8pt}
\end{table}


\section{Limitation and Future Work}
In our experiments, the principal challenge we addressed is the agent's ability to handle varying viewpoints. 
However, real-world recognition scenarios often present a broader and more complex array of challenges, such as varying lighting conditions. 
We leave the investigation for our future work.


\section{Conclusion}
\label{sec:conclusion}
This paper is driven by dual motivations: firstly, to enhance the capabilities of active recognition agents in handling open vocabulary using recent CLIP models, and secondly, to overcome the inherent limitations of CLIP in unconstrained embodied perception scenarios. 
We begin by conducting a quantitative evaluation of CLIP models, assessing their performance across varying viewpoints and occlusion levels. 
This evaluation reveals a marked sensitivity to suboptimal viewing conditions. 
To counteract this, we introduce an active open-vocabulary recognition agent designed to intelligently acquire informative visual observations. 
A critical innovation in our approach is the integration of a self-attention module, which selectively weighs different frames to maintain essential information within the global feature. 
Additionally, we employ similarity measures to avoid the introduction of class-related biases, thereby enhancing the agent's robustness to novel classes. 
Our empirical studies, accompanied with ablation analyses, validate the effectiveness of the proposed agent on both datasets.


{
    \small
    \bibliographystyle{ieeenat_fullname}
    \bibliography{main}
}


\end{document}